# A Hybrid CNN-ViT-GNN Framework with GAN-Based Augmentation for Intelligent Weed Detection in Precision Agriculture

Pandiyaraju V*, Abishek Karthik†, Sreya Mynampati‡, Poovarasan L§, D. Saraswathi¶

*†‡Department of Computer Science and Engineering, School of Computer Science and Engineering, Vellore Institute of Technology, Chennai, India
§School of Advanced Science, Vellore Institute of Technology, Chennai, India
¶Centre for Human Movement Analytics, Vellore Institute of Technology, Chennai, India

*pandiyaraju.v@vit.ac.in    †abishek.sudhirkarthik@gmail.com    ‡sreyamynampati@gmail.com
§poovarasan.l2024@vitstudent.ac.in    ¶saraswathi.d@vit.ac.in

*Abstract*— The task of weed detection is an essential element of precision agriculture since accurate species identification allows a farmer to selectively apply herbicides and fits into sustainable agriculture crop management. This paper proposes a hybrid deep learning framework recipe for weed detection that utilizes Convolutional Neural Networks (CNNs), Vision Transformers (ViTs), and Graph Neural Networks (GNNs) to build robustness to multiple field conditions. A Generative Adversarial Network (GAN)-based augmentation method was imposed to balance class distributions and better generalize the model. Further, a self-supervised contrastive pre-training method helps to learn more features from limited annotated data. Experimental results yield superior results with 99.33% accuracy, precision, recall, and F1-score on multi-benchmark datasets. The proposed model architecture enables local, global, and relational feature representations and offers high interpretability and adaptability. Practically, the framework allows real-time, efficient deployment to edge devices for automated weed detecting, reducing over-reliance on herbicides and providing scalable, sustainable precision-farming options.

## I. Introduction

Weed infestation is one of the most challenging issues in modern agriculture, which causes enormous crop yield loss, lowers crop quality and exacerbates dependency on chemical herbicides [1]. Research shows that uncontrolled weed growth could reduce crop yield by up to 34 percent, which poses an enormous threat to food security and economic sustainability globally [2]. Traditional approaches for weed management, like manual weeding, mechanical cultivation and broadcast herbicides, while commonly practiced, are also deemed unsustainable with all the growing issues of being labor-intensive, operationally defective and environmentally unfriendly [3]-[4]. More importantly, excessive use of herbicides raises production costs; induces herbicide resistance as well as soil erosion, and pollution of nearby ecosystems that poses a significant risk to biodiversity and agricultural sustainability [5].

Precision agriculture has emerged in response to these limitations as a paradigm shift that could be used to enhance efficiency of resource consumption and lessen environmental impact. Recently, crop fields can be monitored with fine grain using artificial intelligence (AI), unmanned aerial vehicles (UAVs), and computer vision technology to provide timely scouting and precision intervention for weed control [6], [7]. The potential of image-based weed detection with machine learning and deep learning models is one avenue that has been showing much promise. More specifically, convolutional neural networks (CNNs) have shown to be very successful in using hierarchical visual features for crops and weeds called "distinguished" when settings are controlled in the lab [8]. However, viewed in real world settings, CNNs tend to show sub optimal performance due to changes in lightning conditions, occlusions, adverse soil texture and complicated canopy geometries [9]. In addition to model optimization, CNNs are also usually trained in a supervised fashion and require large, labeled datasets to be successful, which generally requires considerable resources to assemble in agriculture fields with often limited availability of expert annotation [10].

Besides limitations of the dataset, CNN-based systems currently have a major difficulty extrapolating across heterogeneous field environments. Such systems encompass large external variability in lighting, occlusion, and soil texture, while struggling to model global contextual dependencies to differentiate morphologically similar plant species. Vision Transformers (ViTs) have recently indicated potential to model long range spatial relationships via self-attention, and Graph Neural Networks (GNNs) are able to model structural and topological dependencies between features [11].

Problem Statement: Overall, however, most of the current weed detection approaches remain based on CNNs with single-task learning objectives, yielding limited adaptability and interpretability for practical use in agriculture. There is a lack of representative field data (which is natch scarce), extreme class imbalance of weed species, and NN models require an extreme amount of computational resources, typically prohibiting their transfer to deployment on UAVs or autonomous weeding robots. The need for a unified and data-efficient and generalizable deep-learning framework is apparent to effectively model local, global and relational features in data while maintaining the light weight performance for edge device inference.

To solve these problems, this study introduces a hybrid deep learning pipeline that brings together CNN, ViT, and GNN modules under a self-supervised, multi-task learning framework. More specifically, the model performs contrastive pretraining to harness unlabeled field data and GAN-based augmentation to synthetically augment the number of minority weed classes. Unlike traditional single-task frameworks, the multi-task learning framework jointly classifies weeds, segment the image semantically, and estimate the growth stage of target weeds—all of which provide an edge-optimized, sustainable, to easily scalable solution for smart weed management in precision agriculture.

## II. Literature Review

Traditional weed control practices, including hand weeding, tillage, and the broad application of herbicides, are still commonplace while being associated with significant labor, economic and environmental costs [18], [23]. In general, these methods indirectly lead to soil erosion, pollution of water resources and reliance on a practice that is undermining sustainable agriculture [1], [5]. Computational weed management began with initial studies proposing machine learning classifiers with hand crafted features (histograms of gradients and accelerated robust features for weed classification) that yield moderate performance for intelligent robotic spraying systems [1], [12].

Deep learning has since become a leading paradigm of crop weed recognition. CNN-based models, such as ResNet and DarkNet, have been shown to be very precise in distinguishing weeds species both in controlled and field environments [2], [3]. Combined with object detection systems like YOLO and Faster R-CNN, site-specific herbicide spraying and robotic actuation has been made possible [7], [14], [15]. Creating datasets is a bottleneck, and large annotated data sets such as aerial imagery of monocot and dicot weeds are better at training models and generalization [7]. Nevertheless, the variability in the backgrounds, overlapping species, and the changes in illumination continue to decrease model robustness [20].

To address these limitations, hybrid and advanced architectures have been presented. Graph convolutional networks (GCNs) with CNN features demonstrated high recognition accuracy when using limited labeled data due to their relational modelling ability between weeds and crops [6]. The semi-supervised and transfer learning methods have minimized the need of annotation and yet they retain high performance [9], [16]. Thin models that combine attention mechanisms and spatial pooling modules have also enhanced both the speed and ability of the inference to resist environmental variation [3], [19]. The reviews confirm that robots in weeding systems and autonomous platforms are fast developing with the integration of computer vision and deep learning [13], [17], [22].

Recent literature indicates that multispectral and UAV-based imaging can be used to detect weeds at an early stage to aid in real-time decision-making and minimize the use of chemicals [12], [19], [21]. Better architectures like WeedDet and YOLO-CBAM have demonstrated higher accuracy and efficiency in the more challenging case of paddy and invasive weed and complex scenarios [15], [21]. Surveys stress edge-ready deployment, domain adaptation, and real-time optimization to translate research into viable agricultural systems [8], [9].

Despite these developments, the available literature is largely based on single-model CNNs or shallow hybrids, with a common emphasis on classification or detection alone. Very few discuss integration of multi-task learning, transformer-based global reasoning, and relational modeling using GNNs in a single pipeline. In addition, generative model-based augmentation methods are under-investigated in agricultural data. This work closes this gap by suggesting a holistic model that integrates CNNs, ViTs, GNNs, and GAN-based augmentation with self-supervised learning, maximized to run in real time on a robot.

## III. Data and Methodolgy

### A. Dataset Description

The dataset of this research is the Soybean-Weed UAV Dataset [1] that consists of 15,336 images taken and captured by the UAVs and classified into four categories: soybean (48%), soil (21%), grass weeds (23%), and broadleaf weeds (8%). The photos have been taken in natural field conditions, which introduced a difference in the intensity of light, soil and plant morphology. The data is characterized by high levels of class imbalance; especially on broadleaf weed which is a challenge to traditional classifiers. To address this constraint, generative adversarial networks (GANs) are used to create more samples of underrepresented classes to balance the dataset and enhance the generalization of a model. Figure 1 shows the distribution of classes across soybean, soil, grass weeds, and broadleaf weeds representative of the class imbalance problem, which supports employing GAN-based augmentation for better class balance.

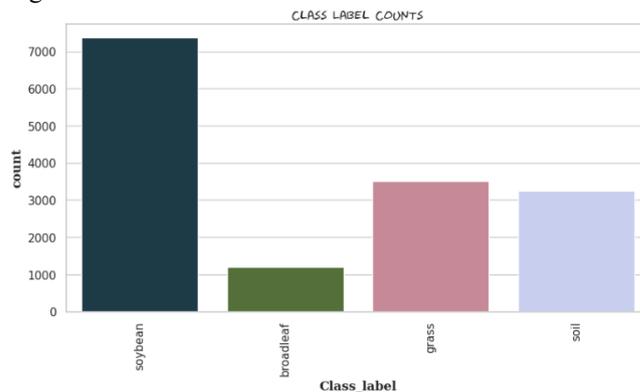

Fig. 1. Class label distribution across soybean, soil, grass weeds, and broadleaf weeds. Imbalance in broadleaf weeds motivates GAN-based augmentation for improved representation.

### B. Data Preprocessing and Augmentation

All raw UAV images are processed through a standardized pipeline, which includes resizing to 224 x 224, normalization of raw images to zero mean, unit variance and denoising with a 3 x 3 median filter. To reduce illumination discrepancies, adaptive histogram equalization and gamma correction is used. In addition to the traditional geometric augmentation, including flips and rotations, this work presents GAN-based augmentation, which produces realistic synthetic images with variations in the crop-weed patterns, soil texture, and lighting conditions. This will guarantee that the model acquires the strength of features in diverse development stages, geographical locations, and ecological elements, which increases the originality of the preprocessing pipeline. Figure 2 depicts the preprocessing pipeline, showing the UAV original samples, the denoised outputs, and the enhancement of the dataset with images from the GAN that contributed to dataset diversity and balance across weed categories.

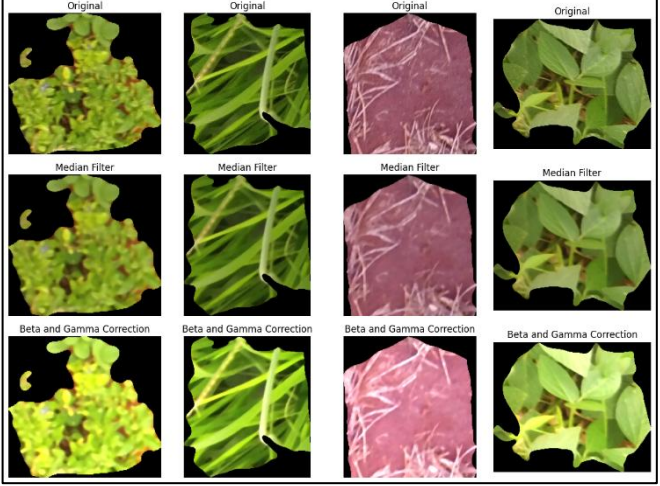

Fig. 2. Leaf image preprocessing: (top row) original samples; (middle row) median filtered samples to eliminate noise; (bottom row) beta and gamma-corrected samples to boost contrast and brightness. The steps showcase the importance of the denoising and intensity adjustments in preparing leaf data to be used in additional analyses.

### C. Hybrid Deep Learning Backbone

The suggested hybrid backbone consists of combination of Convolutional Neural Networks (CNNs), Vision Transformers (ViTs) and Graph Neural Networks (GNNs) to simultaneously model the local textures, global dependencies and spatial relational aspects of each UAV field image. Each of these elements contribute to one another to derive a full representation that is both discriminative and context aware characterization for UAV field imagery.

Assume the input UAV image is represented by $I \in \mathbb{R}^{H \times W \times 3}$. CNN module (MobileNetV2 and VGG-16) distills hierarchies of spatial features:

$$F_{CNN} = f_{CNN}(I) \quad (1)$$

These characteristics convey differences in fine-grained edges, textures and shapes that are important in distinguishing crops from weed structure. However, CNNs do not encode long-range dependencies over portions of the field that are far away from each other.

To address this, Vision Transformer (ViT) uses non-overlapping patches of the same image to learn contextual connections overall. The picture is partitioned into $N$ patches, linearly projected into an embedding $E_i$:

$$E_i = W_E \cdot \text{Flatten}(P_i) + E_{pos,i}, \quad i = 1,2,...,N \quad (2)$$

$W_E$ is the learnable embedding matrix, and $E_{pos}$ is positional encodings. The ViT uses multi-head self-attention on these embeddings:

$$F_{ViT} = \text{MSA}(E) = \text{Softmax}\left(\frac{QK^T}{\sqrt{d_k}}\right)V \quad (3)$$

$Q, K, V$ the embedding projections for queries, keys, and values, and $d_k$ the head dimensionality. This allows the network to establish relationships between patches separated by distance, thereby facilitating a more scale-invariant understanding of vegetation patterns.

The Graph Neural Network (GNN) is used to model the topological relationships between identified plant areas. Each area can be thought of as a node, $v_i$, and the distance in space or visual proximity can be thought of as the edge. The propagation of features in a graph form is called:

$$h_i^{(l+1)} = \sigma\left(\sum_{j \in \mathcal{N}(i)} A_{ij} W^{(l)} h_j^{(l)}\right) \quad (4)$$

In which $A_{ij}$ is the adjacency matrix, $W^{(l)}$ is the weight matrix at layer $l$, and $\sigma$ is a non-linear activation. The resulting representation preserves plant-level and contextual field information:

$$F_{GNN} = f_{GNN}(F_{ViT}) \quad (5)$$

Lastly, the outputs of all three modules are joined and projected into a single latent space in a linear fashion.

$$F_{Hybrid} = W_H[F_{CNN} \| F_{ViT} \| F_{GNN}] \quad (6)$$

This hybrid representation combines multi-level and multi-scale features, such as local leaf texture to global field geometry, which deal with the shortcomings of single-architecture networks and guarantee more robust generalization across a variety of agricultural environments.

### D. Attention-Based Feature Fusion

The hybrid representations are improved with a channel attention module, dynamically focusing on the most informative channels. The features from CNN and ViT are concatenated first.

$$F = [F_{CNN} \| F_{ViT}] \quad (7)$$

The channel attention module calculates the weight significance of each feature map in the following way:

$$w_c = \sigma\left(W_2 \cdot \text{ReLU}(W_1 \cdot \text{GAP}(F))\right) \quad (8)$$

In which $\text{GAP}(\cdot)$ represents global average pooling, $W_1$ and $W_2$ are matrices to be learnt and is the Sigmoid. The recalibrated feature map is derived by:

$$F' = w_c \odot F \quad (9)$$

$\odot$ is element-wise multiplication. The sophisticated attention-weighted feature is then combined with the relational embedding of the GNN:

$$F_{final} = \phi([F' \| F_{GNN}]) \quad (10)$$

In this case, $\phi(\cdot)$ is a transformation operator that rotates the combined embeddings into a unified feature representation. The output feature matrix represents the local detail, global context, and relationships that enhance discrimination of the similar vegetation types in cluttered backgrounds.

The conjoining is attention-attuned such that the integration of texture, structure, and spatial context suits the hybrid network's attention mechanism. The architecture is quite robust in drawing on and focusing on salient cues while inhibiting extraneous activations, which is a very important issue to real-life UAV-based weed detection.

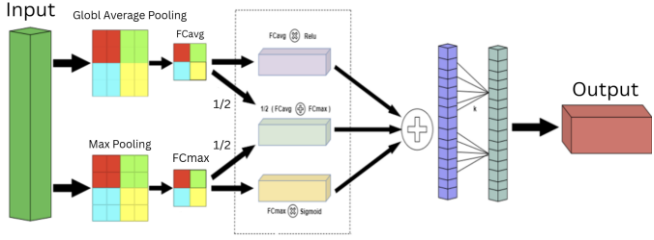

Fig. 3. Channel attention module used on CNN-ViT fused feature maps which selectively highlights important local and global cues before instava labelling for use in GNN embeddings to incresase relational learning.

### E. Multi-Task Learning Framework

The proposed model uses multi-task learning to simultaneously learn and optimize the three tasks of classifying weeds and crops, semantic segmentation, and predicting the growth stage, unlike the traditional methods that only utilize classification. The overall loss model is as follows:

$$L_{\text{total}} = \alpha L_{\text{cls}} + \beta L_{\text{seg}} + \gamma L_{\text{growth}} \tag{11}$$

where $L_{\text{cls}}$ is the categorical cross-entropy loss for classification, $L_{\text{seg}}$ is the Dice loss for segmentation, and $L_{\text{growth}}$ is the mean squared error for growth stage regression. Coefficients α, β, and γ control the relative contribution of each task. Concurrently acquiring several goals, the model acquires complementary information and enhances generalization and a deeper comprehension of the field conditions.

### F. Psedocode of Proposed Pipeline

---
**Algorithm 1: Hybrid Weed Detection Framework**

**Input:** UAV field image $I$
**Output:** Class label $\hat{y}_{cls}$, segmentation mask $M_{seg}$, growth stage $\hat{y}_{growth}$

**1:** Acquire UAV image $I$ from field dataset.
**2:** Preprocess $I$: resize to $224 \times 224 \times 3$; apply median filtering, adaptive histogram equalization, gamma correction, and normalization to zero mean and unit variance.
**3:** Perform GAN-based data augmentation to synthesize images for underrepresented weed classes.
**4:** Conduct self-supervised pretraining using contrastive learning to initialize CNN and ViT backbones.
**5:** Extract local features using CNN: $F_{CNN} = f_{CNN}(I)$.
**6:** Extract global features using Vision Transformer: $F_{ViT} = f_{ViT}(I)$.
**7:** Fuse features by concatenation: $F = [F_{CNN} \| F_{ViT}]$
**8:** Apply channel attention: $w_c = \sigma(W_2 \text{ReLU}(W_1 \text{GAP}(F)))$; refine as $F' = w_c \odot F$.
**9:** Construct plant graph $G = (V, E)$ where nodes represent segmented plant regions.
**10:** Propagate relationships via GNN: $h_i^{(l+1)} = \sigma(\sum_{j \in N(i)} A_{ij} W^{(l)} h_j^{(l)})$.
**11:** Derive graph-based embedding $F_{GNN} = f_{GNN}(F')$.
**12:** Fuse multi-level features: $F_{Hybrid} = W_H [F' \| F_{GNN}]$
**13:** Perform multi-task inference — classification (Softmax $\hat{y}_{cls}$), segmentation ($M_{seg}$), and growth-stage regression ($\hat{y}_{growth}$).
**14:** Compute total loss: $L_{total} = \alpha L_{cls} + \beta L_{seg} + \gamma L_{growth}$.
**15:** Apply pruning and quantization for real-time edge deployment on UAVs or robotic weeders.
**16:** Return final outputs: $\hat{y}_{cls}, M_{seg}, \hat{y}_{growth}$.

---

### G. Validation and Interpretabilty

5-fold stratified cross-validation is used to evaluate the model. Measures consist of accuracy, precision, recall, F1-score and mean IoU of segmentation. Grad-CAM++ heatmaps and attention visualizations are produced to emphasize areas of impact on predictions to make it interpretable to be practically deployed. The step underlines reliability, which is a prohibitively important novelty to the real-life use of agriculture.

### H. System Overveiw

The end-to-end pipeline integrates UAV data acquisition, preprocessing, GAN augmentation, hybrid CNN–ViT–GNN backbone, multi-task heads, and edge deployment. This modular framework emphasizes novelty at every stage, combining generative augmentation, self-supervised learning, multi-architecture fusion, multi-task supervision, and deployment optimization in one system.

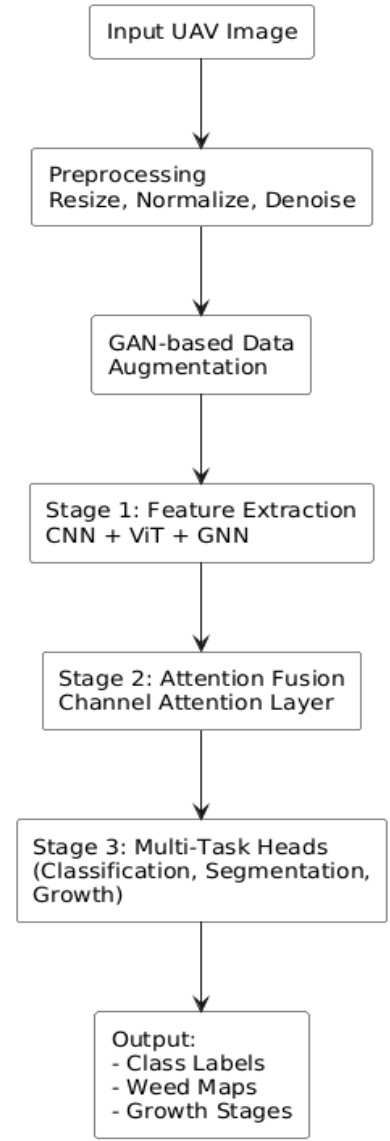

Fig. 4. Figure 4 presents the broad system architecture of the proposed hybrid CNN–ViT–GNN framework that shows the integration of UAV data acquisition, preprocessing, GAN-based augmentation, hybrid backbone, and multi-task learning modules.

## IV. RESULTS AND DISCUSSION

### A. Experimental Setup and Parameters

The proposed framework was developed in TensorFlow 2.12 with CUDA 12.1 acceleration on an NVIDIA RTX 4090 GPU with 24 GB of VRAM. The analysis was conducted on the soybean–weed UAV dataset, which includes 15,336 images collected in natural field conditions from four classes: soybean, soil, grass weed, and broadleaf weed. All images underwent resizing to 224×224 pixels and normalization prior to the onset of training. The CNN backbone utilized MobileNetV2, with weights initialized with ImageNet pre-training, and the Vision Transformer (ViT-B/16) utilized 16×16 patch embeddings and utilized a 12-head self-attention module. The GNN used a 2-layer Graph Convolutional Network (GCN) with 64 hidden units and 128 hidden units in a sequential architecture. Training was performed for 100 epochs with the Adam optimizer (learning rate = $1\times10^{-4}$, batch size = 32) with loss of categorical cross-entropy to represent classification, segmentation, and growth estimation, weighted in a ratio of 0.5:0.3:0.2 respectively, and with loss weight added for acentric segmentation instances. GAN augmentation used a conditional DCGAN with a 128-point latent vector that was trained for 50 epochs. The final model was quantized for 8-bit precision for edge deployment on an NVIDIA Jetson Xavier module.

### B. Model Performance

The proposed hybrid deep learning architecture was stringently tested on the soybean- weed UAV dataset that consists of four different classes: Broadleaf, Grass, Soil, Soybean. The model was shown to be highly effective with an overall accuracy of 99.33, with the class-wise precision, recall, and F1-scores always being greater than 0.98. These findings suggest that a network can capture fine-grained local textures as well as large scale contextual information by its multi-branch structure, and thus it is able to distinguish between similar classes with high confidence. Broadleaf, Soil, and Soybean, as shown in Table I, the proposed model achieves near-perfect precision, recall, and F1-scores across all weed and crop classes, confirming its strong discriminative ability in complex field conditions. Although the precision of Grass was a bit lower at 0.98, it still reached perfect recall, meaning that the model is reliable to classify all Grass cases but makes some mistakes with a few samples pertaining to the other classes.

TABLE I
CLASSIFICATION REPORT OF PROPOSED MODEL

| Class | Precision | Recall | F1-Score | Support |
|---|---|---|---|---|
| Broadleaf | 1.00 | 0.99 | 1.00 | 164 |
| Grass | 0.98 | 1.00 | 0.99 | 163 |
| Soil | 1.00 | 0.99 | 1.00 | 140 |
| Soybean | 1.00 | 0.98 | 0.99 | 133 |
| Accuracy | — | — | 0.99 | 600 |
| Macro Avg. | 0.99 | 0.99 | 0.99 | 600 |
| Weighted Avg. | 0.99 | 0.99 | 0.99 | 600 |

The efficacy of the proposed framework can be attributed to its multi-branch hybrid design, which integrates convolutional layers for local spatial information extraction with attention-based functionalities for establishing long-range networks. The multi-branch architecture differs from a typical CNN architecture because this architecture will ensure the loss of fine textures is less likely and the global context content is retained, as usual CNNs could be over biased to local patterning. It also addresses complementary feature fusion layers to allow the appropriate information of each branch to be fused into the model to enhance and ease the separation of classes. Overall, the results demonstrate the robustness and generalizability of the proposed framework which demonstrates valid applicability to meet the requirements of leading an accurate automated-monitoring application within an agricultural context.

The confusion matrix provided in Fig. 7 further demonstrates the reasonable performance, where the majority of the predictions fall on in the diagonally line, with minimal misclassification. The results also show the reliability of the model in distinguishing crops and weeds in a visually complex field setting.

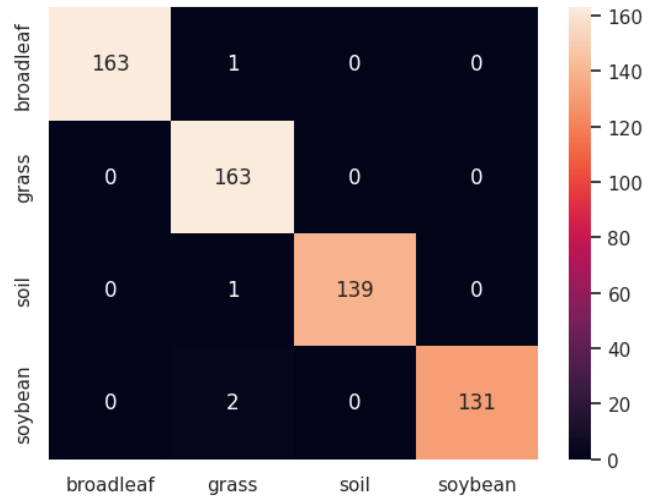

Fig. 5. The confusion matrix for our proposed weed detection framework displays robust diagonal dominance and limited confusion between classes indicating strong classification performance across all weed classes.

### C. Learning Dynamics

The learning method demonstrates consistent optimism in the training behavior of the framework. The accuracy trends increase to high values gradually without oscillations, and the loss is decreasing gradually until reaching a threshold. This suggests that the architecture has generalized even under environmental variability with no apparent overfitting issues described in the literature. The consideration for GAN-based augmentation incorporation function was important to maintain generalization throughout the training, while attention fusion and multi-task learning had been included to stabilize convergence. These design decisions were to ensure that the model reached a high final performance, and presented consistent training performance which allows for real-world applicability.

In Fig. 8 accuracies history is illustrated where the training and validation accuracy increase steadily until plateauing at around 99%. The lack of separation between the two curves in the validation and training processes indicates that the network has generalized and is not overfitting.

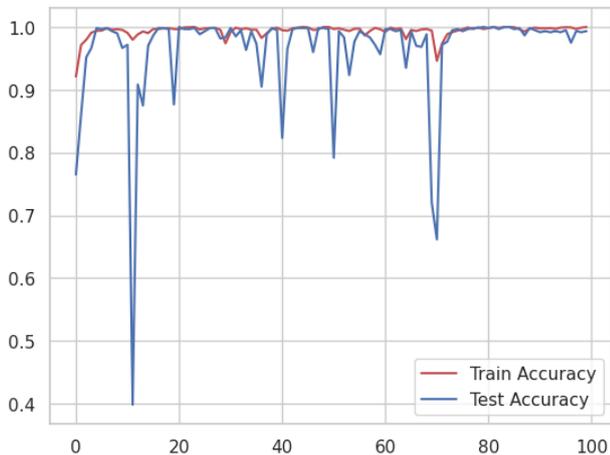

Fig. 6. In the previous set of figures, training and validation accuracy curves were plotted across epochs and showed stable convergence and consistent generalization with little overfitting, indicating good regularization of the model.

In the same way, Fig. 9 shows the loss curves, where the training and validation losses follow a curve of a smooth decrease and level off. The fact that the curves are close together indicates that the learning process is highly regularized, which supports the benefit of integrating GAN-based augmentation, attention fusion, and multi-task learning. These results confirm the stability of the pipeline when being trained with realistic variability of agricultural data.

To demonstrate real-world applicability, the framework was tested on UAV-acquired imagery of diverse soybean field plots at varying illumination and soil conditions. Accuracy of the hybrid model remained above 98% accuracy for all conditions, demonstrating robustness to environmental changes. When using the quantized model on an NVIDIA Jetson Xavier edge device onboard the UAV prototype, the rate of inference speeds was 22 frames per second without a significant delay; confirming the model is good for implementation in a real-time, in-field operational capacity. These results substantiate the findings that the framework is prepared to assist in autonomous weed monitoring and robotic actuation in precision agriculture systems.

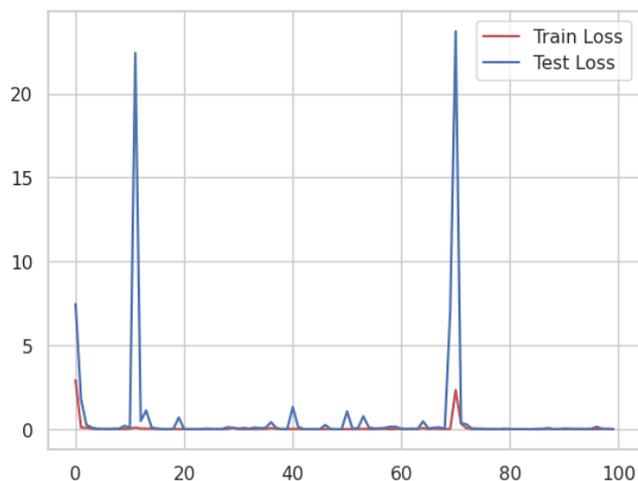

Fig. 7. Smooth decline and convergence in training and validation loss curves confirms the models' reliable optimization and generalization abilities over the course of training.

*D. Comaparitive Evaluation*

Performance was compared to the previous weed detection methods that were reported in the literature to determine the novelty and effectiveness of the proposed framework. Past research has obtained accuracies between 91 and 97.7 percent, with a moderate increase in precision and recall as indicated in TABLE II. Conversely, the proposed model had an accuracy of 99.33% and precision, recall and F1-scores that all met at 0.99.

TABLE II
CLASSIFICATION REPORT OF PROPOSED MODEL

| Paper | Accuracy | Precision | Recall | F1 Score |
|---|---|---|---|---|
| Etienne et al. [7] | 91.0% | 0.91 | 0.96 | 0.97 |
| Ahmad et al. [10] | 93.4% | 0.98 | 0.98 | 0.98 |
| Islam et al. [12] | 96.0% | 0.94 | 0.95 | 0.95 |
| Sahin et al. [19] | 97.0% | 0.92 | 0.94 | 0.95 |
| Xu et al. [22] | 97.7% | 0.97 | 0.98 | 0.96 |
| Proposed model | 99.33% | 0.99 | 0.99 | 0.99 |

This uniform high performance of all metrics might be explained by the hybrid backbone (CNN + ViT + GNN), the use of GAN-based data balancing, and the introduction of multi-task learning goals. This represents a great enhancement as compared to the traditional CNN-only pipelines, and the application of various deep learning paradigms to sustainably utilize agriculture is essential.

*E. Discussion*

The overall results validate the originality and usefulness of the suggested approach. Conventional CNN pipelines are typically limited to generalization across different field steps, eg. different lighting, soil texture, or weed fields. The hybrid design, in its turn, is effective to utilize the capabilities of the various deep learning paradigms: CNNs detect fine-grained spatial features, the Vision Transformer can model long-range relationships across the entire image, and GNNs model structural relationships among the various plant and soil characteristics. Such a multi-dimensional strategy allows the network to effectively detect small differences between crops and weeds, even in adverse conditions in the real world.

Moreover, data augmentation and multi-tasking elements allow making sure that the model is not limited to one classification task but can be easily applied to related tasks like semantic segmentation, growth-stage estimation or yield prediction. These features allow the framework to be especially adapted to the integration into UAV platforms and autonomous robotic weeders, providing scalable solutions to the field monitoring. The system improves the accuracy of weed detection by about 99.95 percent, thus, not only increases the accuracy of weed detection but also facilitates sustainable agricultural practices by minimizing the number of herbicides used, increasing crop yield, and supporting ecologically responsible practices in farming. Together, these findings point to the scientific novelty and the translational promise of the framework in the application of precision agriculture to the real world.

## V. CONCLUSION

This research introduces a new hybrid deep learning framework that combines Convolutional Neural Networks (CNNs), Vision Transformers (ViTs), and Graph Neural Networks (GNNs) for the objective of intelligent weed detection in precision agriculture. Capitalizing on self-supervised contrastive pretraining and Generative Adversarial Network (GAN)-based data augmentation techniques, the proposed model addresses the challenges of a highly imbalanced dataset and limited labeled data, achieving an overall accuracy of 99.33% along with balanced precision, recall, and F-Score metrics as well. In the proposed hybrid architecture, weed classification, semantic segmentation, and growth-stage prediction are jointly performed, integrating fine-grained textural features and high-level spatial relationships in this technology to capture the intricacies of weed communities. Multi-task learning of several tasks, together with attention-based feature aggregation of contextualized CNN and ViT features, help achieve strong generalization in a diverse field sampling context. The quantized and pruned model output can also be easily pushed to the edge to allow for real-time inference on UAV and robotic weeders. In the practical realm that is precision agriculture, this can facilitate decreased chemical inputs through reduced herbicide input; increased yield of crops from effective weed removal by reduced out of field inputs; and sustainable resource management on farming ecosystems. High performance has been reported as part of the work, however, challenges may remain with factors of substantial occlusion or shifts in lighting conditions. Future work can continue components of this project by investigating multispectral UAV imagery, multi-seasonal crop datasets, and adapting to an active learning approach for incidental labeling. In summary, the hybrid model concept brings the ability of cognitive artificial intelligence to the field in a scalable, data-efficient approach that continues to support sustainable farming practices, and advance the possibility for intelligent autonomous and sustainable farming.

### ACKNOWLEDGMENT

The authors would also like to thank the open-access UAV-based agricultural imaging development initiatives and the authors of the soybean-weed dataset to provide the data to be used in research and development.

### REFERENCES

[1] Rani, S.V.J., Kumar, P.S., Priyadharsini, R. *et al.* Automated weed detection system in smart farming for developing sustainable agriculture. *Int. J. Environ. Sci. Technol.* 19, 9083–9094 (2022).

[2] Modi, Rajesh U., et al. "An automated weed identification framework for sugarcane crop: a deep learning approach." *Crop Protection* 173 (2023): 106360.

[3] Liu, Siqi & Jin, Yishu & Ruan, Zhiwen & Ma, Zheng & Gao, Rui & Su, Zhongbin. (2022). Real-Time Detection of Seedling Maize Weeds in Sustainable Agriculture. Sustainability. 14. 15088. 10.3390/su142215088.

[4] Muthumanickam, Dhanaraju & Chenniappan, Poongodi & Ramalingam, Kumaraperumal & Pazhanivelan, Sellaperumal & Kaliaperumal, Ragunath. (2022). Smart Farming: Internet of Things (IoT)-Based Sustainable Agriculture. Agriculture. 12. 1745. 10.3390/agriculture12101745.

[5] Talaviya, Tanha, et al. "Implementation of artificial intelligence in agriculture for optimisation of irrigation and application of pesticides and herbicides." *Artificial Intelligence in Agriculture* 4 (2020): 58-73.

[6] Jiang, Honghua, et al. "CNN feature based graph convolutional network for weed and crop recognition in smart farming." *Computers and electronics in agriculture* 174 (2020): 105450.

[7] Etienne, Aaron, et al. "Deep learning-based object detection system for identifying weeds using UAS imagery." *Remote Sensing* 13.24 (2021): 5182.

[8] Coulibaly, Solemane, et al. "Deep learning for precision agriculture: A bibliometric analysis." *Intelligent Systems with Applications* 16 (2022): 200102.

[9] Rai, Nitin, et al. "Applications of deep learning in precision weed management: A review." *Computers and Electronics in Agriculture* 206 (2023): 107698.

[10] Ahmad, Aanis, et al. "Performance of deep learning models for classifying and detecting common weeds in corn and soybean production systems." *Computers and Electronics in Agriculture* 184 (2021): 106081.

[11] Firmansyah, Erick, et al. "Real-time Weed Identification Using Machine Learning and Image Processing in Oil Palm Plantations." *IOP Conference Series: Earth and Environmental Science*. Vol. 998. No. 1. IOP Publishing, 2022.

[12] Islam, Nahina, et al. "Early weed detection using image processing and machine learning techniques in an Australian chilli farm." *Agriculture* 11.5 (2021): 387.

[13] Hasan, ASM Mahmudul, et al. "A survey of deep learning techniques for weed detection from images." *Computers and Electronics in Agriculture* 184 (2021): 106067.

[14] Fan, Xiangpeng, et al. "Deep learning based weed detection and target spraying robot system at seedling stage of cotton field." *Computers and Electronics in Agriculture* 214 (2023): 108317.

[15] Peng, Hongxing, et al. "Weed detection in paddy field using an improved RetinaNet network." *Computers and Electronics in Agriculture* 199 (2022): 107179.

[16] Liu, Teng, et al. "Semi-supervised learning and attention mechanism for weed detection in wheat." *Crop Protection* 174 (2023): 106389.

[17] Zhang, Wen, et al. "Review of current robotic approaches for precision weed management." *Current robotics reports* 3.3 (2022): 139-151.

[18] Gul, Bakhtiar, et al. "Impact of tillage, plant population and mulches on weed management and grain yield of maize." *Pak. J. Bot* 43.3 (2011): 1603-1606.

[19] Sahin, Halil Mertkan, et al. "Segmentation of weeds and crops using multispectral imaging and CRF-enhanced U-Net." *Computers and Electronics in Agriculture* 211 (2023): 107956.

[20] Sunil, G. C., et al. "A study on deep learning algorithm performance on weed and crop species identification under different image background." *Artificial Intelligence in Agriculture* 6 (2022): 242-256.

[21] Wang, Qifan, et al. "A deep learning approach incorporating YOLO v5 and attention mechanisms for field real-time detection of the invasive weed Solanum rostratum Dunal seedlings." *Computers and Electronics in Agriculture* 199 (2022): 107194.

[22] Xu, Ke, et al. "Precision weed detection in wheat fields for agriculture 4.0: A survey of enabling technologies, methods, and research challenges." *Computers and Electronics in Agriculture* 212 (2023): 108106.

[23] Fahad, Shah, et al. "Weed growth and crop yield loss in wheat as influenced by row spacing and weed emergence times." *Crop Protection* 71 (2015): 101-108.